\pdfoutput=1

\documentclass[11pt]{article}

\usepackage{ACL2023}

\usepackage{times}
\usepackage{latexsym}
\usepackage{booktabs}
\usepackage[T1]{fontenc}

\usepackage[utf8]{inputenc}

\usepackage{microtype}
\usepackage{multirow}
\usepackage{graphicx}
\usepackage{enumitem}
\usepackage{fancyhdr}
\usepackage{inconsolata}
\usepackage{pgfplots}
\pgfplotsset{compat=1.17}

\title{MasonTigers@LT-EDI-2024: An Ensemble Approach Towards Detecting Homophobia and Transphobia in Social Media Comments}

\author{Dhiman Goswami, Sadiya Sayara Chowdhury Puspo, Md Nishat Raihan,\\ \textbf{Al Nahian Bin Emran} \\
        George Mason University, USA \\
        \texttt{\{dgoswam, spuspo, mraihan2, abinemra\}@gmu.edu} \\
        }

\begin{document}
\maketitle
\begin{abstract}
In this paper, we describe our approaches and results for Task 2 of the LT-EDI 2024\footnote{\url{https://codalab.lisn.upsaclay.fr/competitions/16056}} Workshop, aimed at detecting homophobia and/or transphobia across ten languages. Our methodologies include monolingual transformers and ensemble methods, capitalizing on the strengths of each to enhance the performance of the models. The ensemble models worked well, placing our team, \textit{MasonTigers}, in the top five for eight of the ten languages, as measured by the macro F1 score. Our work emphasizes the efficacy of ensemble methods in multilingual scenarios, addressing the complexities of language-specific tasks.
\end{abstract}

\section{Introduction}
In this current era dominated by social media platforms, people heavily rely on online content for communication, learning, knowledge-sharing, and staying abreast of new technologies. The comment sections, intended for constructive feedback, unfortunately, sometimes become grounds for hate speech, offensive comments, and discrimination, including targeting a specific community. Such behaviors cause trauma, fear, anxiety, depressive symptoms and discomforts among LGBTQ+ individuals \citep{poteat2014short, ventriglio2021homophobia}, hindering them from freely expressing their thoughts and feedback.

To ensure the safety and comfort of users on online platforms, it becomes imperative to identify and address hate speech and offensive comments. Although there are existing policies aimed at protecting communities from such misconduct, violations may lead to the removal of offending comments \footnote{\url{https://www.youtube.com/howyoutubeworks/our-commitments/standing-up-to-hate/}} \footnote{\url{https://transparency.fb.com/policies/community-standards/hate-speech/}}. However, the identification process necessitates the application of NLP and AI techniques due to the diverse nature of hate speech, which can manifest in both direct and passive forms. Surprisingly, there is a higher focus on researching this topic in English and other high-resource languages like hindi, with ample resources. For instance, couple of shared tasks have been organized previously e.g. \citet{chakravarthi-etal-2022-overview}, \citet{chakravarthi2022overview}, \citet{chakravarthi-homophbia-2023-overview}, accompanied by substantial datasets e.g. \citet{vasquez2023homo}, \citet{chakravarthi2021dataset}. However, there has been a lack when it comes to identifying hate speech in low-resource and under-resource languages.

This shared task \citet{homophobia-2024-overview} aims to identify hate speech contents, specifically homophobia, transphobia, and non-anti-LGBT+ sentiments, directed at LGBTQ+ individuals in 10 different languages, including 7 low-resource languages. To tackle the linguistic diversity, we conduct separate experiments for each language, leveraging different transformer-based models with proficiency in distinct languages. Notably, for Tulu, an under-resourced language, we employ a prompting approach. Alongside these experiments, various techniques are explored, and the most effective one during the evaluation phase is implemented in the test phase for comprehensive validation.

\section{Related Works}
As smart devices, mobile apps, and social media platforms become more widely utilized, there are also more negative effects linked to them, such as cyberbullying, hurtful comments, and rumors have increased. These platforms have also become a space for negative behaviors like sexism, homophobia (\citet{Sarah}), misogyny (\citet{mulki2021letmi}), racism (\citet{larimore2021}) and transphobia (\citet{giametta2021mapping}). Internet trolling, where people say mean things online, has become a global problem. To deal with this, researchers are looking into automated methods since checking every message manually is impossible. In this section, we provide a brief summary of the research attempts that focused on identifying homophobia, transphobia, and non-anti-LGBT+ content from YouTube comments.

To motivate researchers to tackle the issue of identifying toxic language directed at the LGBTQ community, in the last few years, several shared tasks have been released. In one such shared task related to LT-EDI-2022 (\citet{chakravarthi2022}), researchers submitted systems to deal with homophobic and transphobic comments. \citet{chakravarthi-etal-2022-overview} gives an overview of the models submitted. Three subtasks for the languages Tamil, English, and Tamil-English (code-mixed) were the emphasis of this shared task. 
Apart from this, several studies have been conducted. The author of \citet{Karayigit2022HomophobicAH} used the M-BERT model, which shows that it is capable of accurately identifying homophobic or other abusive language in Turkish social media comments. Similarly, another shared task related to this topic has been organized showing XLM-R performing best with spatio-temporal data in 5 different languages \citep{wong2023cantnlp}. \citet{vasquez2023homo} presents a mexican-spanish annotated corpus along in which beto-cased (Spanish BERT) outperforms the other models. 

For several text classification tasks, transformer-based ensemble approaches perform very well, like the works by \citet{goswami2023nlpbdpatriots, raihan2023nlpbdpatriots}. Also, prompting Large Language Models like GPT3.5 \cite{OpenAI2023GPT4TR} is another popular approach in recent classification tasks \cite{raihan2023offensive} for the past year.

Efforts to identify homophobic and transphobic comments have primarily focused on a maximum of five languages to date. However, in this shared task, a total of 10 languages have been chosen, and ongoing efforts now include Telugu, Tulu, and Marathi.

\begin{figure}[!b]
\centering
\scalebox{.92}{
\begin{tikzpicture}[node distance=1cm]
    \tikzstyle{block} = [rectangle, draw, fill=blue!20, text width=\linewidth, text centered, rounded corners, minimum height=4em]
    \tikzstyle{operation} = [text centered, minimum height=1em]
    \node [block] (rect1) {\textbf{Role:}{ "You are a helpful AI assistant. You are given the task of detecting homophobia and transphobia in a given text. }};
    
    \node [operation, below of=rect1] (plus1) {};
    
    \node [block, below of=plus1] (rect2) {\textbf{Definition:}{ Homophobia and transphobia detection is the process of identifying expressions of hatred or discrimination against LGBTQ+ individuals in communication.'. }};

    \node [operation, below of=rect2] (plus2) { };
    
    \node [block, below of=plus2] (rect3) {\textbf{Examples:}{ An example of Homophobic/Transphobic comment: \textbf{<Example1>}. An example of Non-Homophobic/Transphobic comment: \textbf{<Example2>}'. }};
    
    \node [operation, below of=rect3] (plus3) {};
    
    \node [block, below of=plus3] (rect4) {\textbf{Task:}{ Generate the label [YES/NO] for this \textbf{"text"} in the following format: \textit{<label> Your\_Predicted\_Label <$\backslash$label>}. Thanks."}};
    
\end{tikzpicture}
}
\caption{Sample GPT-3.5 prompt for few shot learning [\textit{Used for the \textbf{Tulu} Dataset}].}
\label{fig:prompt1}
\end{figure}

\section{Datasets}
The dataset provided for the shared task contains 10 languages - Tamil, English, Malayalam, Marathi, Spanish, Hindi, Telugu, Kannada, Gujarati, and Tulu. It is compiled using five separate research works. The previous iteration of the workshop \cite{chakravarthi-homophbia-2023-overview} includes Tamil, English and Spanish, Hindi, Malayalam languages, and the earlier version \citet{Kumaresan-homophobia-2024-overview} includes Tamil, English, and Tamil-English (Code-Mixed) languages.

The work by \citet{kumaresan2023homophobia} builds a dataset for Malayalam and Hindi languages from social media comments. Another dataset by \citet{chakravarthi2023detection} focuses on YouTube comments. One data augmentation approach is adopted by \citet{chakravarthi2022can}. 

The current dataset combines all these works to build a comprehensive dataset for the task of homophobia and/or transphobia detection in 10 languages \citet{homophobia-2024-overview}.

The detailed dataset demonstration and label-wise data percentage for all the languages are available in Table \ref{tab:label}.

\begin{table*} [!t]
\centering
\scalebox{0.95}{
\begin{tabular}{lccc|lccc}
\hline
\multicolumn{4}{|c|}{\textbf{Tamil}} & \multicolumn{4}{c|}{\textbf{English}} \\
\hline
Labels & Train & Dev & Test & Labels & Train & Dev & Test \\
\hline
Non-anti-LGBT+ content & 77.53 & 76.13 & 76.11 & Non-anti-LGBT+ content & 94.12 & 94.45 & 94.04\\
Homophobia & 17.02 & 17.72 & 18.25 & Homophobia & 5.66 & 5.30 & 5.56\\
Transphobia & 5.45 & 6.15 & 5.64 & Transphobia & 0.22 & 0.25 & 0.40\\
\hline
\multicolumn{4}{|c|}{\textbf{Malayalam}} & \multicolumn{4}{c|}{\textbf{Marathi}} \\
\hline
Labels & Train & Dev & Test & Labels & Train & Dev & Test \\
\hline
Non-anti-LGBT+ content & 79.25 & 77.25 & 77.83 & None of the categories & 73.49 & 72.13 & 75.87\\
Homophobia & 15.29 & 16.24 & 16.17 & Homophobia & 15.74 & 17.20 & 14.93\\
Transphobia & 5.46 & 6.51 & 6.00 & Transphobia & 10.77 & 10.67 & 9.20\\
\hline
\multicolumn{4}{|c|}{\textbf{Spanish}} & \multicolumn{4}{c|}{\textbf{Hindi}} \\
\hline
Labels & Train & Dev & Test & Labels & Train & Dev & Test \\
\hline
None & 58.34 & 51.82 & 50.00 & Non-anti-LGBT+ content & 94.65 & 95.31 & 95.95\\
Transphobic & 20.83 & 24.09 & 25.00 & Transphobia & 3.59 & 4.06 & 3.12\\
Homophobic & 20.83 & 24.09 & 25.00 & Homophobia & 1.76 & 0.63 & 0.93\\
\hline
\multicolumn{4}{|c|}{\textbf{Telugu}} & \multicolumn{4}{c|}{\textbf{Kannada}} \\
\hline
Labels & Train & Dev & Test & Labels & Train & Dev & Test \\
\hline
None of the categories & 38.63 & 38.51 & 38.37 & None of the categories & 44.35 & 44.27 & 44.11\\
Homophobia & 32.12 & 31.18 & 32.18 & Homophobia & 28.17 & 28.61 & 28.11\\
Transphobia & 29.25 & 30.31 & 29.45 & Transphobia & 27.48 & 27.12 & 27.78\\
\hline
\multicolumn{4}{|c|}{\textbf{Gujarati}} & \multicolumn{4}{c|}{\textbf{Tulu}} \\
\hline
Labels & Train & Dev & Test & Labels & Train & Dev & Test \\
\hline
None of the categories & 47.39 & 45.29 & 45.63 & NON H/T & 74.25 &  & 82.32\\
Homophobia & 27.93 & 28.62 & 29.31 & H/T & 25.75 &  & 17.68\\
Transphobia & 24.68 & 26.09 & 25.06 &  &  &  &\\
\hline
\end{tabular}}
\caption{Label-wise Data Percentage for Different Languages}
\label{tab:label}
\end{table*}

\begin{table*} [!h]
\centering
\begin{tabular}{lcc|lcc}
\hline
\multicolumn{3}{|c|}{\textbf{Tamil (Rank 5)}} & \multicolumn{3}{c|}{\textbf{English (Rank 10)}} \\
\hline
Models & Dev F1 & Test F1 & Models & Dev F1 & Test F1 \\
\hline
XLM-R & 0.49 & 0.49 & XLM-R & 0.32 & 0.32 \\
mBERT & 0.62 & 0.67 & mBERT & 0.32 & 0.32 \\
TamilBERT & 0.51 & 0.53 & roBERTa & 0.32 & 0.32 \\
\hline
Wt. (Dev F1) Ensemble  & & 0.51 & Wt. (Dev F1) Ensemble  & & 0.32 \\
Wt. (Test F1) Ensemble  & & \textbf{0.52} & Wt. (Test F1) Ensemble  & & 0.32 \\
\hline
\multicolumn{3}{|c|}{\textbf{Malayalam (Rank 9)}} & \multicolumn{3}{c|}{\textbf{Marathi (Rank 4)}} \\
\hline
Models & Dev F1 & Test F1 & Models & Dev F1 & Test F1 \\
\hline
XLM-R & 0.51 & 0.51 & XLM-R & 0.49 & 0.49 \\
mBERT & 0.54 & 0.55 & mBERT & 0.46 & 0.41 \\
MalayalamBERT & 0.52 & 0.52 & MarathiBERT & 0.41 & 0.44 \\
\hline
Wt. (Dev F1) Ensemble  & & 0.51 & Wt. (Dev F1) Ensemble  & & 0.44 \\
Wt. (Test F1) Ensemble  & & \textbf{0.52} & Wt. (Test F1) Ensemble  & & \textbf{0.45} \\
\hline
\multicolumn{3}{|c|}{\textbf{Spanish (Rank 3)}} & \multicolumn{3}{c|}{\textbf{Hindi (Rank 4)}} \\
\hline
Models & Dev F1 & Test F1 & Models & Dev F1 & Test F1 \\
\hline
XLM-R & 0.75 & 0.55 & XLM-R & 0.33 & 0.33 \\
mBERT & 0.79 & 0.55 & mBERT & 0.33 & 0.33 \\
SpanishBERT & 0.81 & 0.50 & HindiBERT & 0.33 & 0.33 \\
\hline
Wt. (Dev F1) Ensemble  & & 0.50 & Wt. (Dev F1) Ensemble  & & 0.33 \\
Wt. (Test F1) Ensemble  & & \textbf{0.54} & Wt. (Test F1) Ensemble  & & 0.33 \\
\hline
\multicolumn{3}{|c|}{\textbf{Telugu (Rank 1)}} & \multicolumn{3}{c|}{\textbf{Kannada (Rank 3)}} \\
\hline
Models & Dev F1 & Test F1 & Models & Dev F1 & Test F1 \\
\hline
XLM-R & 0.97 & 0.97 & XLM-R & 0.95 & 0.95 \\
mBERT & 0.96 & 0.95 & mBERT & 0.93 & 0.94 \\
TeluguBERT & 0.97 & 0.97 & KannadaBERT & 0.95 & 0.95 \\
\hline
Wt. (Dev F1) Ensemble  & & 0.97 & Wt. (Dev F1) Ensemble  & & 0.95 \\
Wt. (Test F1) Ensemble  & & 0.97 & Wt. (Test F1) Ensemble  & & 0.95 \\
\hline
\multicolumn{3}{|c|}{\textbf{Gujarati (Rank 5)}} & \multicolumn{3}{c|}{\textbf{Tulu (Rank 3)}} \\
\hline
Models & Dev F1 & Test F1 & Models & Dev F1 & Test F1 \\
\hline
XLM-R & 0.94 & 0.94 & GPT 3.5 & & 0.45 \\
\cline{4-6}
mBERT & 0.95 & 0.95 & XLM-R & 0.42 & 0.42 \\
GujaratiBERT & 0.93 & 0.93 & mBERT & 0.42 & 0.45 \\
\cline{1-3}
Wt. (Dev F1) Ensemble  & & 0.94 & KannadaBERT & 0.42 & 0.45 \\
\cline{4-6}
Wt. (Test F1) Ensemble  & & 0.94 & Wt. (Test F1) Ensemble & & 0.45 \\
\hline
\end{tabular}
\caption{Combined Results for Various Languages}
\label{tab:combined_results}
\end{table*}

\section{Experiments}

Among the 10 languages (except Tulu) we use XLM-R, m-BERT and language specific BERTs: RoBERTa \cite{DBLP:journals/corr/abs-1907-11692} , HindiBERT \cite{nick_doiron_2023},  TamilBERT \cite{joshi2023l3cubehindbert}, MalayalamBERT \cite{joshi2023l3cubehindbert}, MarathiBERT \cite{joshi2022}, SpanishBERT \cite{canete-etal-2022-albeto}, TeluguBERT \cite{joshi2023l3cubehindbert}, KannadaBERT \cite{joshi2023l3cubehindbert}, GujaratiBERT \cite{joshi2023l3cubehindbert} for Hindi, Tamil, Malayalam, Marathi, Spanish, Telugu, Kannada, Gujarati respectively.  While using MarathiBERT and HindiBERT, we pad the sentence length upto 512 tokens, because of the limitation of the aforementioned BERTs. Training parameters are mostly kept the same across all models, mentioned in Table \ref{tab:training_config}.

After that we perform weighted ensemble approach of the above models (XLM-R, m-BERT, language specific BERTs) and use the macro F1 scores of the models on the dev data as the weight along with the confidence score to get the ensemble macro F-1 score of the test phase. After the test labels get published, we use the F1 score in the rank list as the weight along with the corresponding confidence score for the additional experiment of weighted ensemble approach. 

\begin{table}[ht]
\centering
\begin{tabular}{lc}
\toprule
\textbf{Parameter} & \textbf{Value} \\
\midrule
Learning Rate & \(1e-5\) \\
Train Batch Size & 8 \\
Test Batch Size & 8 \\
Epochs & 5 \\
\bottomrule
\end{tabular}
\caption{Training Configuration Parameters}
\label{tab:training_config}
\end{table}

As Tulu is very close to Kannada and Tulu doesn't have any language specific fine-tuned model, we used KannadaBERT on Tulu. In South Karnataka, individuals who speak Tulu are typically fluent in both Tulu and Kannada. Due to the long-standing interaction between Tulu and Kannada, it can be anticipated that codeswitching between these two languages is a probable outcome \citep{shetty2003language}. Moreover, we implement few shot learning using GPT3.5 for the Tulu dataset (see Figure \ref{fig:prompt1}). Such prompting is very widely used in recent works on text classification \citep{raihan2023sentmix, goswami2023offmix}. We got the same result as the ensemble approach in the few shot prompting technique. We specifically used 8 - shot prompting for Tulu language. This process is inspired by \citet{wei2022chain}.

We improve the macro F1 in 4 out of 10 cases in this additional experiments and rest of the 6 cases we get the same macro F1 as the test phase.

\section{Results}

We employ an ensemble-based methodology for the tasks, since in text classification tasks this can further improve the results. for For all the ensemble approaches, we use XLM-R, mBERT, and a BERT-based model fine-tuned for that specific langauge as we mentioned in the previous section.

In the testing phase, we ensemble the confidence score of the three models and then calculate the weighted average. The weight in this context is the macro F1 score of the corresponding models on dev data.

For Tulu language, we use GPT 3.5 for few shot prompting. We use few instances of each of the two labels with specific prompt and and then get the test labels predicted by GPT 3.5.

After the end of testing phase when the test labels get published, we further run the ensemble systems again for all the languages. For this case, the weight for the models is the macro F1 score of the corresponding models on test data. For Tulu language, we also peform this ensemble approach with XLM-R, mBERT and KannadaBERT. Though none of these models have Tulu language in their corpora but all these models were pretrained on Kannada which is really close to Tulu. We achieve a macro F1 score using this ensemble approach which is as same as the few shot prompting.

Using this approach, we achieve the rank 1 (one) on Telugu, Rank 3 (three) on Spanish, Kannada and Tulu, rank 4 (four) on Marathi and Hindi, rank 5 (five) on Tamil and Gujarati, rank 9 (nine) on Malayalam and rank 10 (ten) on English language. The detailed experimental results of the models are available in Table \ref{tab:combined_results}.

\section{Error Analysis}

To thoroughly investigate the results and the models' performance on specific datasets, we find that though the accuracy of the models on all the datasets are very good but the macro F1 score is really low in some cases. From the table in Table \ref{tab:label}, it is clearly visible that English and Hindi dataset is very imbalanced. They have a very few Homophobia and Transphobia label. From the confusion matrices in Appendix \ref{sec:appendix2}, we can see all the instances of Non-anti-LGBT+ content label are correctly predicted by the models but models' fail to be well-trained on the other two labels. Thus the other two labels get mis-classified which leads to a marco F1 score around 0.33 for these two languages. For Tamil, Malayalam, Marathi, Spanish and Tulu - the data ratio is comparatively balanced which leads to F1 score in the range 0.45 - 0.54. Telugu, Kannada and Gujarati datasets are evenly label-wise evenly balanced which get reflected with highest marco F1 score in the range 0.94 - 0.97. For the imbalanced datasets, widely used techniques like data augmentation, back translation can be proven helpful which can be potential future research scope in this domain. Detailed error analysis for the languages is given below:
\begin{itemize}

\item{\textbf{Tamil}} The Homophobia instances are partially correct but all the instances of Transphobia are misclassified. On the other hand, the Non-anti-LGBTQ + content instances are almost perfectly classified in Tamil.

\item{\textbf{English}}
The models perform well in identifying only Non-anti-LGBT+ content while they completely fail to detect Homophobia, and Transphobia in English.

\item{\textbf{Malayalam}}
For Malayalam, the models almost perfectly detect Non-anti-LGBTQ + Content, partially detect Homophobia instances but as before it completely misclassified the Transphobia. 

\item{\textbf{Marathi}}
The Homophobia and  None of the categories instances are partially correct in case of Marathi. However, the transphobia content instances are mostly misclassified.

\item{\textbf{Spanish}}
Homophobic, None and Transphobic instances are partially correct in Spanish.

\item{\textbf{Hindi}}
Only Non-anti-LGBTQ + content instances are perfectly classified but all the instances of Homophobia and Transphobia are misclassified in Hindi.

\item{\textbf{Telugu}}
Homophobia, None of the categories and Transphobia instances are almost perfectly correct in Telugu.

\item{\textbf{Kannada}}
The model predictions are almost correct in Kannada for instances of  Homophobia, None of the categories and Transphobia. 

\item{\textbf{Gujarati}}
Our model can almost perfectly detect for Homophobia, None of the categories for Gujarati Transphobia instances.

\item{\textbf{Tulu}}
Though NON H/T instances are perfectly classified, but all the instances of H/T are misclassified in Tulu.
\end{itemize}

\section{Conclusion}

The task of detecting abusive speech targeting sexual and gender minorities has become increasingly important given the rise of social media and its potential to propagate harmful stereotypes that further marginalize vulnerable populations. This paper presents our efforts to address online transphobia and homophobia in multilingual contexts, which remains an under-studied area in abusive language detection. We employ an ensemble approach combining multiple individual models to identify abusive speech across datasets in ten languages.

Our findings demonstrate that while no individual system consistently achieves superior performance across all data, monolingual language-specific BERT models fine-tuned on our abusive speech data unsurprisingly emerge as strong approaches for this classification problem. However, our ensemble framework leveraging voting across multiple BERT variants along with other models surpasses any individual system, indicating the value of model diversity even when one base technique manifests strengths. We hypothesize that the inconsistencies across models and languages result partly from imbalanced, sparse instances of actual abusive samples in our data. Hence, future work should prioritize constructing larger, more balanced benchmark datasets for abusive language detection encompassing underrepresented identities. Nonetheless, this research presents a starting point for identifying multilingual, multidirectional abuse in online spaces through ensemble natural language systems.

\section*{Limitations}
The monolingual BERT models underperformed due to insufficient data volume and class imbalance in our existing abusive speech corpora. Skewed label distributions with far fewer minority abuse cases than benign texts make learning discriminative patterns difficult. Our ensemble framework mitigated these weaknesses but still suffered performance issues on minority samples. Constructing larger, more balanced datasets remains imperative yet challenges exist regarding sensitivity of human annotations for this problem space. Nonetheless, enhancing model robustness on sparse abusive instances should be prioritized. While augmenting through back-translation and generation could help, this risks polluting training data if new variants stray from actual phenomenology. Systems producing false positives that ascribe nonexistent abuse contribute harm. Progress on mitigating imbalance without these downsides is incremental. Our experiments manifested datasets with endemic skew away from minority classes. Researchers must remain cognizant that efforts to populate abusive classes risk drifting from reality. 

\section*{Ethics Statement}
Adhering to the \href{https://www.aclweb.org/portal/content/acl-code-ethics}{ACL Ethics Policy}, this study seeks to responsibly progress online safety through benign content filtering technology. However, safeguards against misuse for censorship/monitoring remain imperative. While the supplied dataset was anonymized to protect privacy, carefully compiled public benchmarks avoiding marginalization must drive future progress. Flawed training data has propagated harm before; our experiments mitigated this but work continues. Guiding principles of beneficence and nonmaleficence should steer research on automating content classification with real-world impacts, including on complex offensive speech. Assessing for unintended consequences and awareness of social dimensions is critical as this work makes initial strides in detecting minority-targeting online abuse. Continual reassessment of systems and their real-world influences remains essential even beyond research contexts. And usage policies must be crafted thoughtfully before any operational deployment. We believe impactful technology like this carries with it a profoundly ethical mandate. Progress ceases to be progress if attained through forfeiture of our values.

\bibliography{anthology,custom}
\bibliographystyle{acl_natbib}

\appendix

\section{Confusion Matrix}

\label{sec:appendix2}

\begin{figure} [!h]
  \centering
  \includegraphics[width=\linewidth]{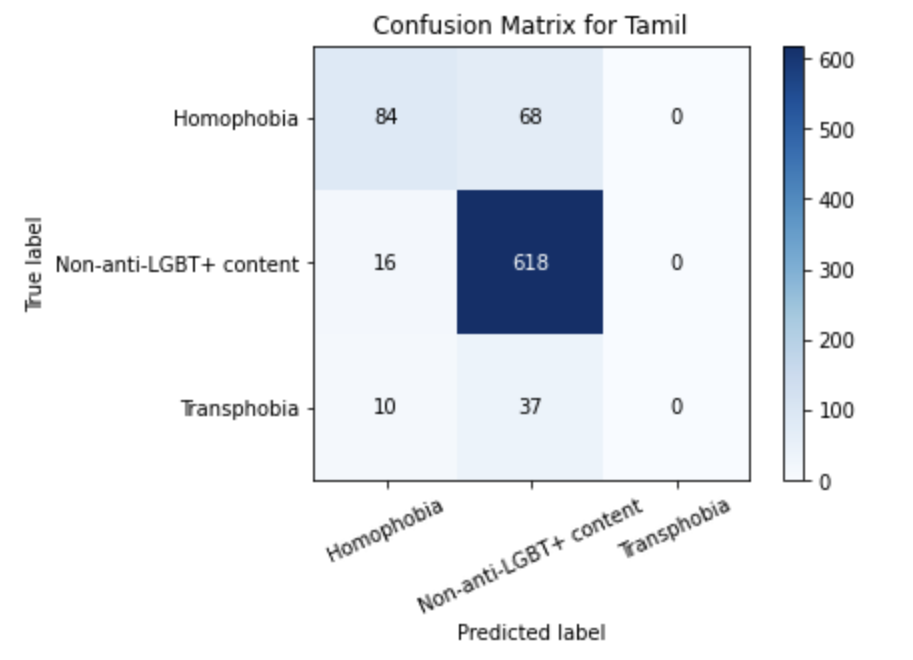}
  \caption{Confusion Matrix for Tamil Language}
  \label{fig:tamil}
\end{figure}

\begin{figure} [!h]
  \centering
  \includegraphics[width=\linewidth]{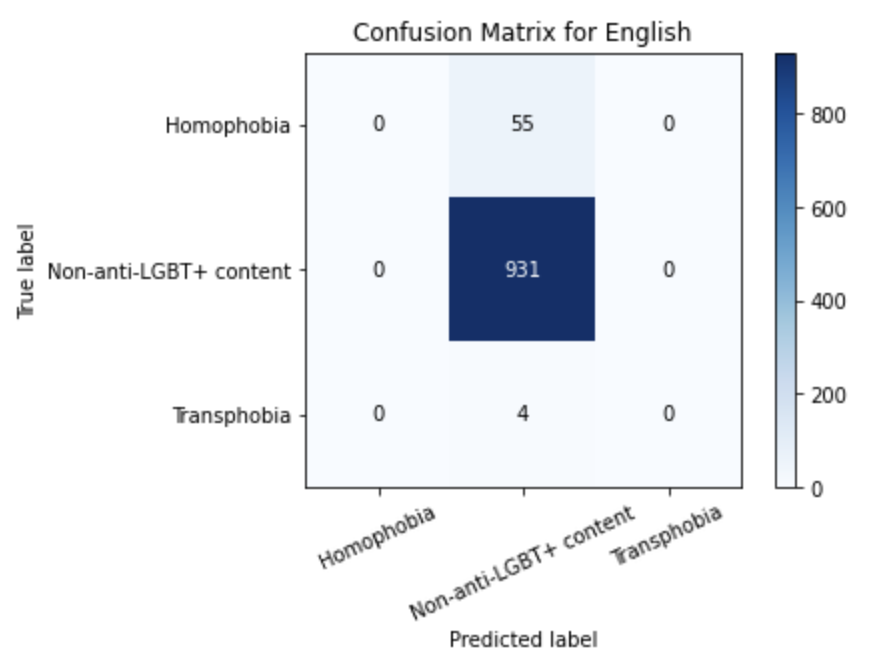}
  \caption{Confusion Matrix for English Language}
  \label{fig:english}
\end{figure}

\begin{figure} [!h]
  \centering
  \includegraphics[width=\linewidth]{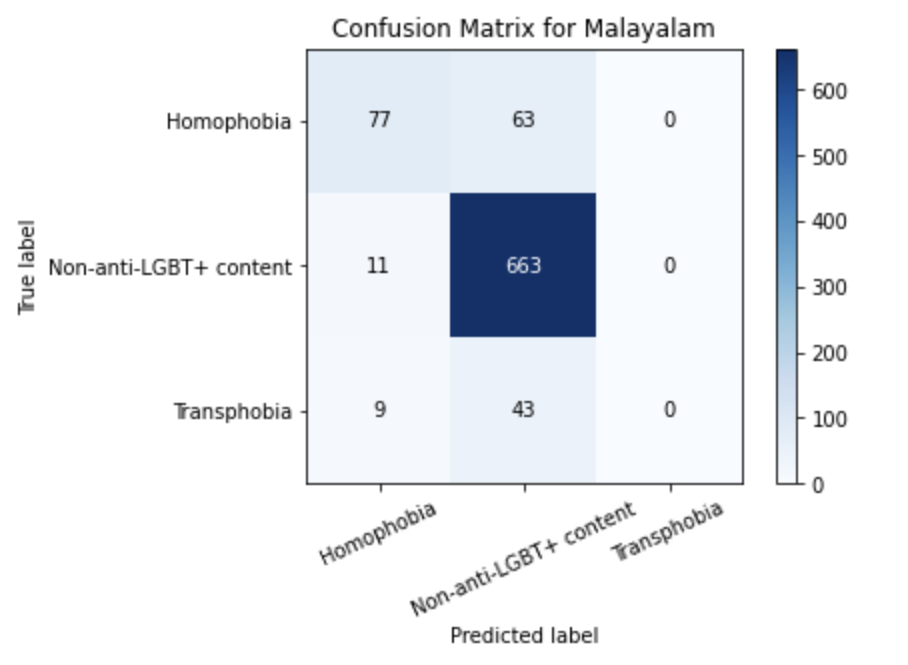}
  \caption{Confusion Matrix for Malayalam Language}
  \label{fig:malayalam}
\end{figure}

\begin{figure} [!h]
  \centering
  \includegraphics[width=\linewidth]{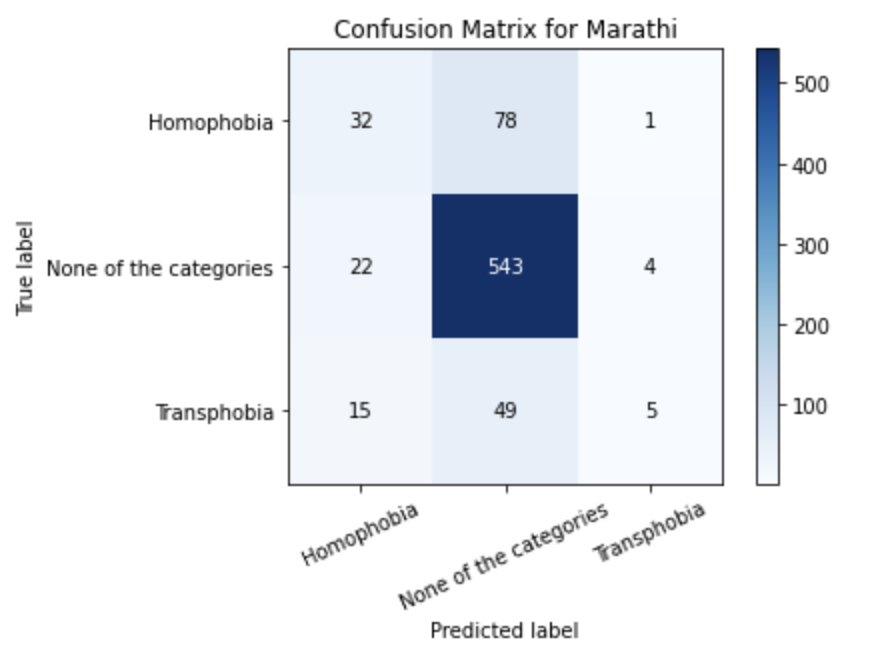}
  \caption{Confusion Matrix for Marathi Language}
  \label{fig:marathi}
\end{figure}

\begin{figure} [!h]
  \centering
  \includegraphics[width=\linewidth]{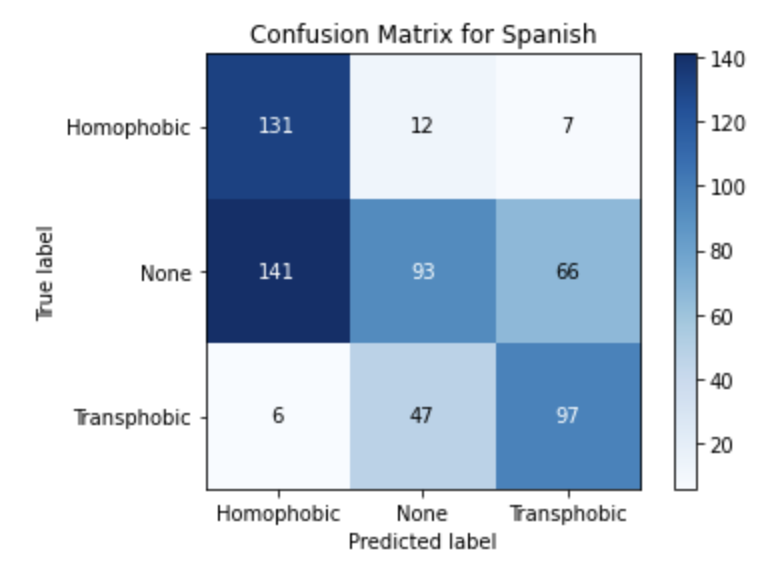}
  \caption{Confusion Matrix for Spanish Language}
  \label{fig:spanish}
\end{figure}

\begin{figure} [!h]
  \centering
  \includegraphics[width=\linewidth]{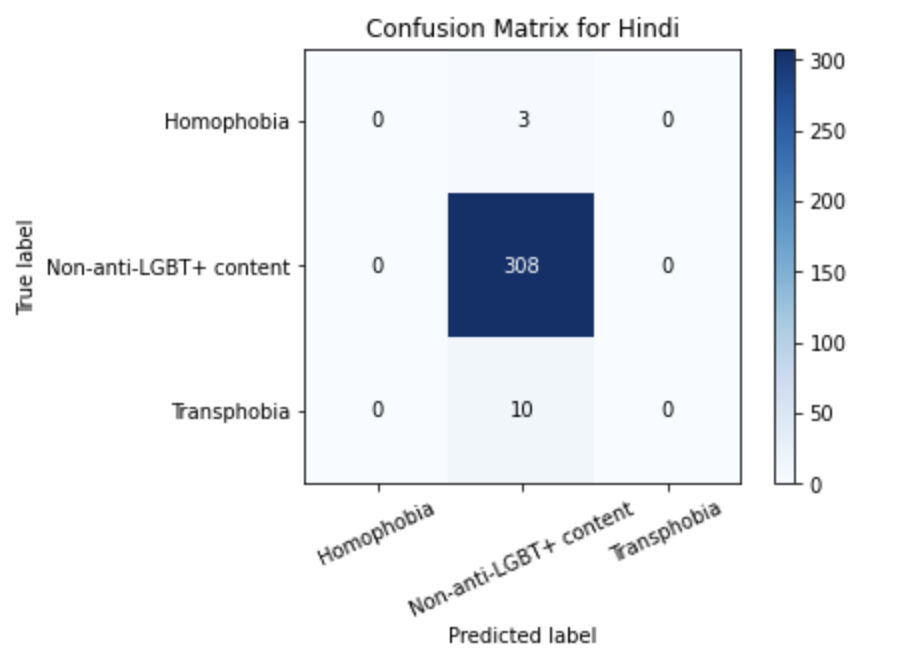}
  \caption{Confusion Matrix for Hindi Language}
  \label{fig:hindi}
\end{figure}

\begin{figure} [!h]
  \centering
  \includegraphics[width=\linewidth]{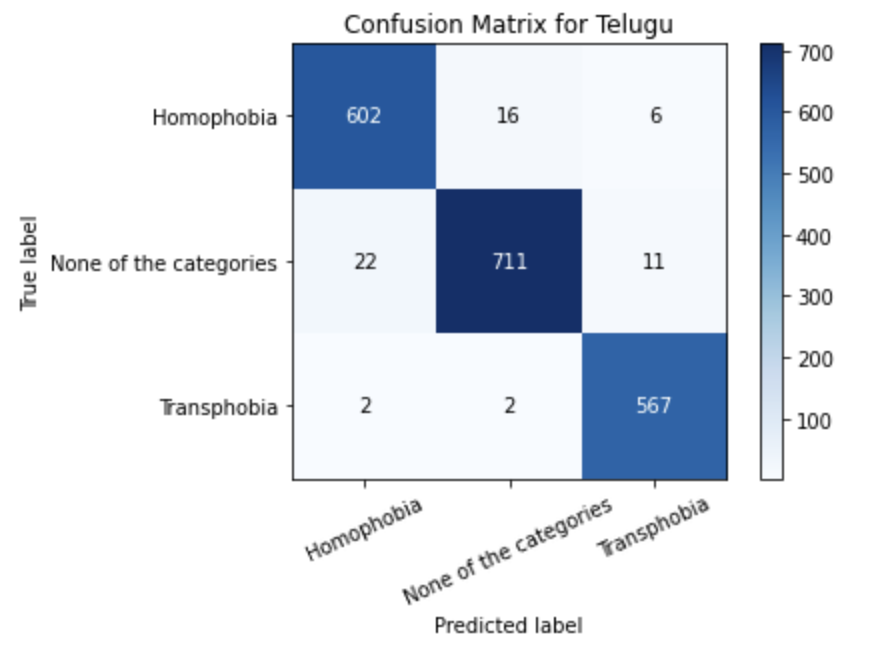}
  \caption{Confusion Matrix for Telugu Language}
  \label{fig:telugu}
\end{figure}

\begin{figure} [!h]
  \centering
  \includegraphics[width=\linewidth]{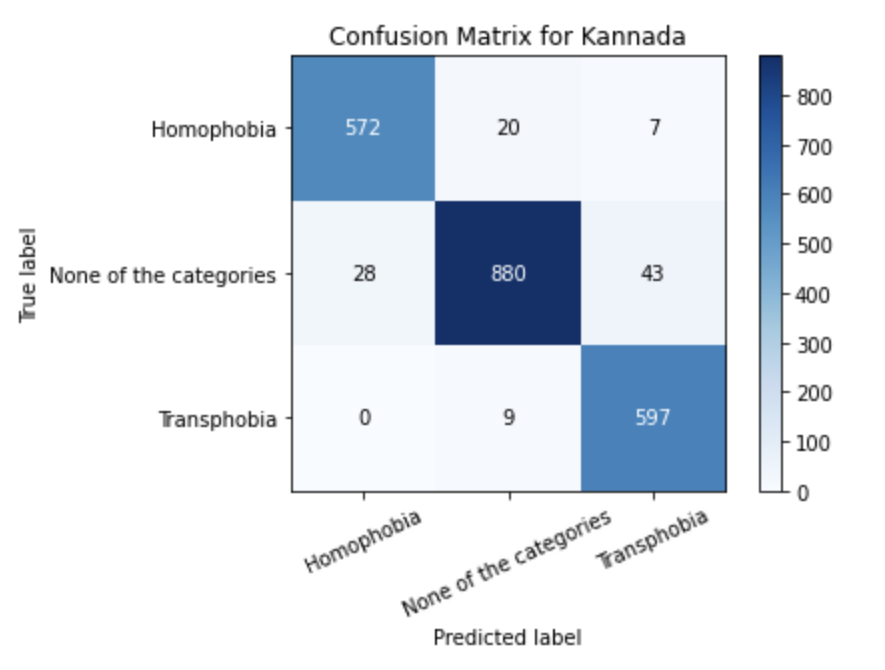}
  \caption{Confusion Matrix for Kannada Language}
  \label{fig:kannada}
\end{figure}

\begin{figure} [!h]
  \centering
  \includegraphics[width=\linewidth]{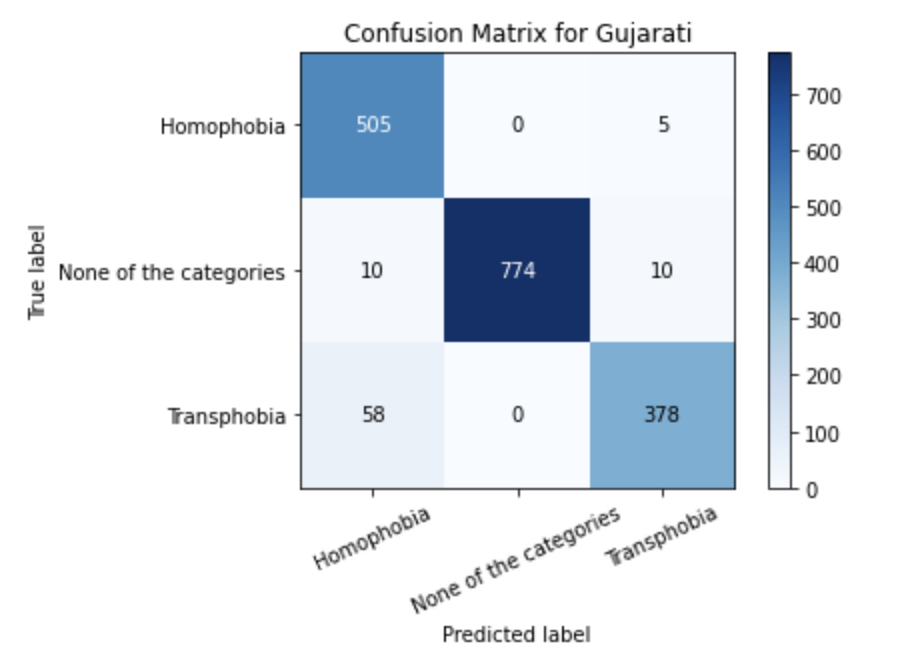}
  \caption{Confusion Matrix for Gujarati Language}
  \label{fig:gujarati}
\end{figure}

\begin{figure} [!h]
  \centering
  \includegraphics[width=\linewidth]{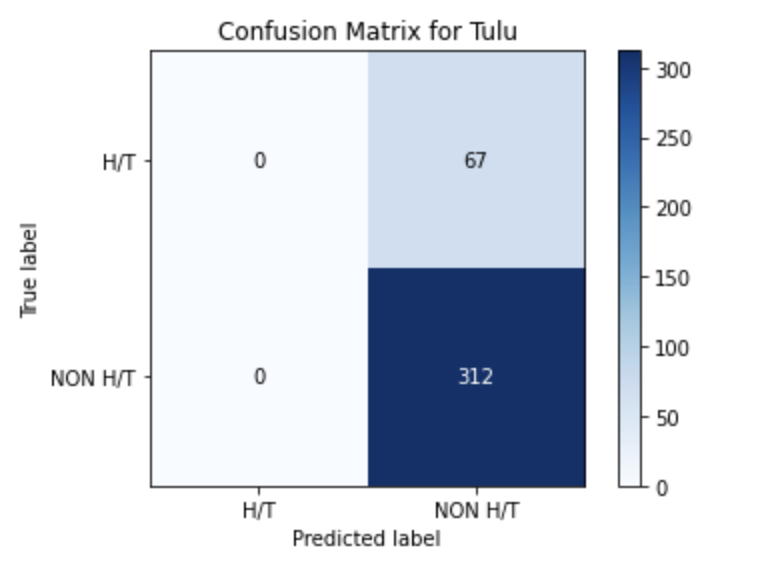}
  \caption{Confusion Matrix for Tulu Language}
  \label{fig:tulu}
\end{figure}

\end{document}